\documentclass{article}

\usepackage{arxiv}

\usepackage[utf8]{inputenc} 
\usepackage[T1]{fontenc}    
\usepackage{hyperref}       
\usepackage{url}            
\usepackage{booktabs}       
\usepackage{amsfonts}       
\usepackage{nicefrac}       
\usepackage{microtype}      
\usepackage{lipsum}
\usepackage{graphicx}
\graphicspath{ {./images/} }

\title{Machine Common Sense}

\author{
 Alexander Gavrilenko \\
  Thinking Technologies\\
  Minsk, Kalinina Str, 1-106\\
  \texttt{mmaud@yandex.by} \\
   \And
 Katerina Morozova \\
  Thinking Technologies\\
  Minsk, Kalinina Str, 1-106\\
  \texttt{katerina.morozova@gmail.com} \\}

\begin{document}
\maketitle
\begin{abstract}
Machine common sense remains a broad, potentially unbounded problem in artificial intelligence (AI). There is a wide range of strategies that can be employed to make progress on this challenge. This article deals with the aspects of modeling commonsense reasoning focusing on such domain as interpersonal interactions.  The basic idea is that there are several types of commonsense reasoning: one is manifested at the logical level of physical actions, the other deals with the understanding of the essence of human-human interactions. Existing approaches, based on formal logic and artificial neural networks, allow for modeling only the first type of common sense. To model the second type, it is vital to understand the motives and rules of human behavior. This model is based on real-life heuristics, i.e., the rules of thumb, developed through knowledge and experience of different generations. Such knowledge base allows for development of an expert system with inference and explanatory mechanisms (commonsense reasoning algorithms and personal models). Algorithms provide tools for a situation analysis, while personal models make it possible to identify personality traits.  The system so designed should perform the function of amplified intelligence for interactions, including human-machine. 
\end{abstract}


\section{Introduction}
Binet and Simon \cite{Bi05} believed that the alteration or the lack of common sense was of the utmost importance for practical life. Irrationality that is due to the lack of common sense can be partly compensated for by the development of intelligence that is defined as ‘a very general mental capability that, among other things, involves the ability to reason, plan, solve problems, think abstractly, comprehend complex ideas, learn quickly and learn from experience’\cite[p.~13]{Go97}. Unfortunately, many people are not able to think abstractly, reason soundly and learn quickly. And, of course, many people strive to find out if there are effective tools to prevent mistakes \cite{Ma15}. 

Therefore, providing a rationale for machine common sense is crucial for AI creation. Much hope is pinned on the intelligence enhancers (amplified intelligence), which can improve application of the rationale knowledge to human behaviour. Finding the solution to such problem, according to Charter and Oaksford \cite{Ch01}, is the major intelligent challenge.

Having considered a proof provided by modern science it becomes obvious that a model for human reasoning is a collection of models of different nature and levels of generalization, one of which shall deal with common sense of interpersonal relations and interaction of human beings in real life, i.e. the spheres where irrational behavior of people and their underdeveloped personality make themselves felt. Unfortunately, we can’t find in the public domain the methodology for modeling commonsense reasoning focusing on such domain as interpersonal interactions. Therefore doing our research we aimed at finding the grounds for the development of a practically applicable method of common sense modeling, which can help people make difficult life decisions. 

In the first part of the study, we discuss different approaches to understanding of common sense. This is important because, as practice shows, the choice of modeling approach and the scope of the model application depends on the understanding of the object.

The second part of the research is devoted to ontological components in expert system able to simulate commonsense reasoning in such domain as interpersonal interactions. These components include: heuristic knowledge base, heuristic algorithms, personal models of users. The knowledge base is focused on heuristics, expressing the common sense of interpersonal interactions. We’ve revealed that a set of heuristics organized in a certain way can be the knowledge base of the expert system able to simulate commonsense reasoning in interpersonal relations. Constraint-based decision-making algorithms with key points in the form of objectively existing constraints are focused on heuristics. User personal models, considering subjective factors that affect the results obtained, are embedded in the algorithms. Such ontology is consistent with the common sense that people use to make decisions, aiming at distinguishing between objective and subjective limitations and finding ways of overcoming them.

The third part deals with practical application of the expert system that simulates the way humans perform commonsense reasoning in human-human interaction. Such a model can: (a) improve the efficiency of decision-making as applied to personal relations, business and politics; (b) assist AI in understanding the situations involving interpersonal interactions;  (c) make voice assistants and personal robots more robust, turn them into consultants helping to solve real-world problems; (d) establish a system in which the machine will teach people rules to reduce the chances of making mistakes; (e) measure potential trust between partners.

\section{Approaches to Development of Machine Common Sense}
\subsection{Bottom-Up and Top-Down Approaches to Modeling}
Having analyzed research practices and scientific literature, we have noticed that the vast majority of common sense research focuses on the bottom-up approach to AI design (Bottom-up AI), involving artificial neural networks and evolutionary computing. \cite{Mc68,Fi71,Re80,Bo84} Such approach instantiates the notion of common sense into physical actions and can be modeled based on training neural networks with big data, including semantics-based formal logic. It is worth noting that semantic concepts are widely used for simulation and modeling of images and speech. Researchers are trying hard to find methods to justify and effectively implement these ideas. \cite{Si02,La15,Tu14,Ra18a,Ra18b} The most notable example of this approach is CYC, a schema of commonsense axioms (facts, empirical semantic rules and heuristics) that have been handcrafted for and entered into CYC's knowledge base. For example, ‘Every tree is a plant’ and ‘Plants finally die’. But, due to lack of clarity in the definition of ‘common sense’, it is not clear which part of common sense is implied in certain areas of CYC. The same problem applies to the models of commonsense reasoning, simulating logic of pure movements and actions, which DARPA and Allen Institute for Artificial Intelligence implement in Machine Common Sense based on artificial neural networks (Bottom-up AI). 

The state-of-the-art results for artificial neural networks in such domains as speech synthesis and images and text recognition and obscure the top-down approach to AI design (Top-Down AI). We mean the development of expert systems. As practice shows, in matters relating to the nature of phenomena and processes, the success of neural networks is not very impressive. According to Chomsky \cite{Ch12}, the bottom-up approach is rather superficial. The fact is that artificial neural networks are nothing like human brains. This is a mathematical apparatus that allows for imitating some principles of brain function. Modern neural networks are successfully applied in speech synthesis and image recognition, they function based on a neural network laws that are far from the actual processes taking place in the human brain.

We have found the most comprehensive overview of their capabilities and limitations at Lake et al \cite{La17}. Let us add some practical considerations.  First, there are research projects at the intersection of biology and mathematics aimed at studying intelligent behavior in animals.\cite{La12,Ch13,An10} Many of them study the evolution of logic presented in mathematical proofs. But mathematical models are unlikely to be a strong reflection of reality, the responses of biological neural networks are logically unpredictable, as they depend not only on electrical but also on complex chemical processes. \cite{An15} We cannot even be sure of the same response to the same stimuli. Secondly, neural networks are expensive to train to solve serious problems. Thirdly, today, there is an objective need in understanding the results obtained with the help of neural networks. All these make us believe the top-down approach to AI design is worth attention.  

The problems of using neural networks for processing abstract knowledge are becoming more and more apparent. \cite{Be19,Co63,God40} Today, there are no reasonable ideas on how to model, based on big data analysis, plausible reasoning in view of interpersonal interactions, including the concepts of interests, common sense and morality. The lack of adequate criteria for personality assessment and the uncertainty of personal perceptions make artificial neural networks ineffective. R. Penrose \cite{Pe94} noted that computation could contribute to understanding, but computation itself provides no actual understanding. What is needed is the idea derived from awareness, the same awareness is required to interpret the calculation results. If we consider the matter closer, it becomes obvious that attempts to displace the element of human awareness and replace it exclusively with computational procedures seem to be unwise, or even impractical. Even in mathematics, computers are useful only in view of top-down calculations. 

As for the programming of commonsense reasoning, it is worth to point out that the first difficulties arise from different understanding of the subject matter of the research.  It is clear that there are several types of commonsense reasoning. One of them manifests itself at the logical level of physical actions; the other implies understanding of complex nature of human-human interaction. Approaches based on formal logic and artificial neural networks allow for modeling only the first type of commonsense reasoning. The fact that physiological processes involved with thinking are for a fraction of a second faster than the transmission of a nerve impulse contributes to the idea. The rules governing the transition from neurophysiological structures to consciousness remain unclear. \cite{Ed01} For second type model (common sense of interpersonal interactions), mathematics alone is not enough, it is vital to understand the motives and rules of human behavior. Human thinking in this case goes beyond the syntactic or mechanical relations, although it is subject to logic (not formal). Davis and Marcus \cite{Da15} draw attention to numerous aspects proving that there are great obstacles to produce a “satisfactory common sense reasoner”.  This is confirmed by the fact that at the moment no research group dealing with neural networks has found the solution to model commonsense reasoning focusing on such domain as interpersonal interactions.

\subsection{Common Sense and Interpersonal Interactions}
Talking about common sense in view of interpersonal interactions one should bear in mind that in real life people in general behave irrationally. Emotions affect decision-making process and make people act regardless of the common sense. Let us consider one illustrative example.  At the 2015 European Games two wrestlers were disqualified for getting into a brawl. The incident occurred at the 57-kilogram freestyle semi-final. At the beginning of the second period, the semi-finalists became involved in a fight with each other which broke up after punches and head-butts were thrown. Both athletes were disqualified. It is obvious that the athletes were affected by emotions and behaved irrationally. The wrestler who was wining succumbed to his emotions, responded and lost, though he could have won both morally and physically, if he left the insult unanswered. 

Kahneman and Tversky \cite{Ka79} documented the manifestation of irrationality in human behavior as a consequence of incorrect perception and information processing and showed how errors of judgment invoked and took roots in the way we think. Later, Selten \cite{Se90} found that the factors that bound the area of rationality referred to motivation and introduced a three-level model of decision making i.e. routine, imagination, and reasoning. One can take different decisions at different levels.  In his opinion, the final decision is not part of rational analysis, it is made solely by the portion of the brain that deals with motivation, not logic.\cite{Se91} Aren’t people capable of making rational decisions? Such assumption contradicts the reality. We agree with Reason \cite{Re90}, who has established the levels at which mistakes occur, i.e. Skill-based Level; Rule-based Level; Knowledge-based Level. It is obvious that cognitive science can’t cope with biological reactions to stimuli. But other mistakes can be avoided due to the knowledge gained. 

The point is that where to acquire commonsense knowledge for decision-making in view of human-human interactions?  The researchers note that most of such knowledge exists in the form of abstract rules transmitted to us through social environment, including friends, parents and teachers \cite{Sc72}. Such abstract rules disclose cause-and-effect relationships between personality traits and outcomes of the actions. Simon \cite[p.~62]{Si19} underlined that in large part human goal-directed behavior simply reflected the shape of the environment in which it took place.  Schedrovitsky \cite[p.~66]{Sc05} supported this point of view and specified that ‘the biggest problems of human thinking are associated with “culture” and its secrets, with our values and meanings that are primarily influenced by social relationships; this, in my opinion, is the most important and most interesting area of human thinking.’ 

Understanding the complex structure of interpersonal interaction contributes to the development of research methods that are based on scenarios and causal reasoning \cite{Sc75}, the study of words of wisdom of prior generations. According to Chater and Oaksford ‘The key to human reasoning appears to lie in how world knowledge provides only the most plausible model of some premises, or accesses only the most relevant information, or permits an assessment of the probabilities of events’ \cite[p.~208]{Ch01}.

Linguists, such as Lakoff and Johnson \cite{La80}, argue that metaphors are the key to understanding the commonsense thought.  Gunning \cite{Gu18}, has a strong position and points out that “Although there is no general agreement on the importance of grounded cognition and metaphor in AI, it seems clear that development of more perceptually grounded representations will be critical for making progress on machine common sense, where matching human concept representations is critical. Such representations would not only get us closer to human cognition, they may also be the key to integrating machine learning and machine reasoning”. Of course, metaphors are not the only source of commonsense knowledge. Moreover, their understanding requires from a person more than zero reflection, as well as metaphor-based modeling demands significant efforts and costs.  There are many other more understandable sources of life wisdom, which are studied by cognitive sciences, the basic of which are philosophy and psychology. \cite{Ga87} As a proof we can cite philosophical ideas of Machiavelli \cite{Ma32}, who described the rules of common sense (real-life heuristics) in terms of retention of power.  His advice was to separate politics from ethics, and substitute force and astuteness for law, paralyze the individual intelligence; to keep concealed from the country what is happening in the world, and likewise from the capital what is happening in the provinces; to weaken public opinion until it subsides in apathy; profit by the ease with which men turn informers; to manipulate society by means of its vices.

Such a policy of the rulers rests on a limited understanding of the common sense and the fear of the people. From the point of view of public interests, fear leads to resistance, causes negative emotions, make people meet the needs elsewhere, it is exemplified by the slave-owning system. Fear requires compensation, which comes in the form of high institutional costs, corruption and fraud.  Latent conflict can manifest itself in poor job performance, hiding or ignoring information, postponing the execution of resolutions, the diversion of resources. Therefore, common sense power retention strategies suggest a compromise between the interests of rulers, power groups and the population. However, lacking high level common sense, many rulers only serve their own interests and they do this as a rule using force and manipulations, especially widespread in the digital era. Then the time comes when the demand for justice increases so much that the existing government may be overthrown. Let us consider the attitude towards pension age as an example of a state having or lacking common sense.  The authorities in Poland and Italy, for instance, reduce the retirement age in public interest. The greatest progress is observed in China, where the retirement age is reduced from 55 to 50 (for women) and from 60 to 55 (for men), while in Russia, the retirement age is unreasonably increased from 55 to 60 (for women) and from 60 to 65 (for men). There are many examples of direct conflict of interest between the parties (events in Venezuela, France, England, etc.), one of which has a low level of common sense.

As evidenced in practice, the higher the level of intelligence of the person is, the more common sense thinking moves him towards understanding the distant consequences and outcomes associated with actions and events. For example, many people overeat. You can say that they act irrationally, against common sense. It is true if viewed from the point of view of a person who is able to assess the effect of such nutrition on health. But those who lack knowledge do not consider the harm done to health, they do not control their interests and desires. In other words, to take informed decisions one needs common sense, i.e. a way to minimize damage to one’s own interests, based on the experience and knowledge gained by a person through living. Therefore, it is reasonable to address machine common sense through the lens of maximum expected utility for the user. ‘In a sense, the MEU principle could be seen as defining all of AI. All an intelligent agent has to do is calculate the various quantities, maximize utility over its actions, and away it goes. But this does not mean that the AI problem is solved by the definition! … Estimating the state of the world requires perception, learning, knowledge representation, and inference. Computing result requires a complete causal model of the world.’ \cite[p.~611]{Ru10}   

The need to consider causal relations between knowledge and actions prioritize the top-down approach (intelligent expert system development) to common sense modeling in view of interpersonal interactions. We can assume that both approaches (bottom-up and top-down) can be used in the creation of the machine common sense.  The bottom-up approach shall be used to model the logic of simple physical processes, while the top-down approach (heuristic engines) shall be employed for high-level abstract reasoning. However, today, as evidenced in practice, the concepts of cognitive science is a big challenge for specialists  involved with artificial neural networks. Russell and Norvig specify that ‘The philosophical literature on minds, brains, and related topics is large and sometimes difficult to read without proper training in the terminology and methods of argument employed.’ \cite[p.~1042]{Ru10} 

\subsection{Heuristic Expert Systems}
Expert systems are as progressive in the field of artificial intelligence as neural networks. At their time, they survived the crisis, the essence of which was formulated by Minsky: ‘Unfortunately, the strategies most popular among AI researchers in the 1980s have come to a dead end. For each different kind of problem, the construction of expert systems had to start all over again, because they didn't accumulate common-sense knowledge.’ \cite{Mi03}

Currently, expert systems are expecting a second birth, because a new technological basis, i.e. artificial neural networks, lacking the content of expert systems, has been developed. The need to search for knowledge and scientific justification of qualitative cause-and-effect dependencies has not disappeared.  According to Russell and Norvig, ‘knowledge-based agents can benefit from knowledge expressed in very general forms, combining and recombining information to suit myriad purposes. They can combine general knowledge with current percepts to infer hidden aspects of the current state prior to selecting actions.’ \cite[p.~194]{Ru03} 

If we are to talk about personal relations, than it should be noted that real life proves that someone who has gained much life experience and knows the way things work is considered to be sound-minded.  In fact, life experience is awareness of a number of real-life heuristics, i.e. common sense rules. This allows us to suggest that a system containing a large number of real-life heuristics may become the basis of the commonsense reasoning model. This corresponds to Heiner’s  research \cite{He83} that has shown that in the context of decision making human beings are guided by ready-made behavior patterns (life heuristics) applicable to most real life situations. He believes that since a set of patterns is limited, heuristic models describe human behavior under uncertainty better than optimization-based ones. Gigerenzer \cite{Gi08} is of the same opinion, he believes that there are situations when a simple heuristic makes more accurate predictions than multiple regression or Bayesian models do. He underlines that from a heuristic model point of view the mind resembles an adaptive toolbox with various heuristics tailored for specific classes of problems much like the hammers and screwdrivers in a handyman’s toolbox. 

We believe that there are enough reasons that determine the fundamental for a heuristic approach to be employed for modeling commonsense reasoning in view of interpersonal interactions. Such possibility is attributable to the factors as follows: 

1. Life situations and diversity of personalities seem to be the world's largest source of primary information. 

2. The number of common sense solutions that minimize people's losses and mistakes is much less than the variety of life situations. 

3. Practical human experience over generations has evolved rules of common sense (real-life heuristics). These rules (life heuristics) are formulated by the wisest representatives of mankind and show the dependence between the outcomes of people's actions and their character traits, including their morality and reasoning skills. Therefore, when it comes to common sense, it is quite reasonable to assume that ‘perhaps the mind is really just a collection of smart heuristics, rather than a fantastically powerful computing machine’. \cite{Ch00} Simon and Newell \cite{Si71} assumed that it is possible to create universal knowledge base from real life heuristics and logically order a list of differences (constraints) between the initial situation and the target situation. Davis and Marcus \cite{Da15} also believe that it is crucial for an expert system, which simulates common sense of interpersonal interactions, to build the knowledge base and a framework to process such knowledge using common sense algorithms.

Heuristic expert systems allow for generating hypotheses that can be used for informed decision-making. The challenge lies in finding a way to model the outcomes of concrete situations using abstract heuristics. We hypothesize that the ontological structure of an expert system that simulates the common sense of interpersonal interactions shall correspond to the logic of reasoning of an individual and the knowledge base of such system shall contain structured heuristics, decision-making algorithms and personal models that are essential to understand certain people. 

\section{Interpersonal Interaction Common Sense Expert System Ontological Structure}
\subsection{Knowledge Database of Expert System}
\subsubsection{Simple Heuristics}
In the scientific literature, heuristics are used as a general approach reflecting some abstract rules. For example, if you can’t find a solution, just imagine that you have found one, and see what you can learn from it. Or, if you have difficulty in understanding the problem, try drawing a picture. \cite{Po45}  Newell, Shaw and Simon \cite{Ne62} concretize the notion of heuristics as a rule that reduces options in choice tasks and consider it an important cognitive tool for decision-making. Todd and Gigerenzer also believe that ‘fast and frugal heuristics employ a minimum of time, knowledge, and computation to make adaptive choices in real environments.’ The advantage of heuristics is that they contain common sense and decision-making experience of many people in an abstract form. \cite[p.~731]{To00} According to Russell and Norvig, ‘were it not for the ability to construct useful abstractions, intelligent agents would be completely swamped by the real world’. \cite[p.~69]{Ru10} The choice of a good abstraction thus involves removing as much detail as possible while retaining validity and ensuring that the abstract actions are easy to carry out. 

We’ve noticed that there is a special class of life heuristics that are used in the relationship between people and life situations.  Simple heuristics with a wide scope are the most studied among them. There are many examples of such heuristics:  Tit-for-tat \cite{Ax84}, Imitate the majority \cite{Bo05}, Imitate the successful \cite{Hu05}, Take-the-best \cite{Ma02}. Meanwhile, much extra data is needed for practical application of simple heuristics. Let us consider as an example the Take-The-Best Heuristics.  To apply this heuristics one should know what the best is. But what are the criteria for the best? What are resource constraints (financial, personal, time) to determine the best? Best for whom?  What is the meaning of the verb ‘to take’: to make, to collect, to buy?  The uncertainty of answers to these questions may result in the heuristic to turn to a call for actions of incomprehensible nature. 

Thus, the main peculiarity of simple heuristics is their prescriptive character.  They do not take into account situational conditions and the personality of the man making decisions. Meanwhile, we have noticed that simple heuristics can be more useful if applied in the complex. Let's take as an example four heuristics, typical of commonsense reasoning

\begin{itemize}
\item Do not lie to yourself!
\item Useful is that what leads to development!
\item Excess is always expensive!
\item Do not provide services you are not asked for!
\end{itemize}

Using common sense of these four rules for decision-making will help to avoid many problems.  Thus, heuristic ‘do not lie to yourself’ refers to the situations when a person’s mistake made in good faith is caused by his/her lack of knowledge.  This heuristic suggests that a person lacking common sense is unable to see the predicted outcomes of the situation as well as the constraints and challenges he/she may face on the way to achieve his/her goal. Therefore, before making a decision, it is vital to acquire the necessary knowledge and get advice. Heuristic ‘useful is that what leads to development’ warns a person that he/she shouldn’t take actions, whose outcomes won’t have positive impact on his/her personality. Heuristic directs thinking for the search of positive criterion of the decisions taken. The application of this heuristic is usually accompanied by overcoming difficulties and suggests the presence of willpower. Heuristic ‘excess is always expensive’ is a strong heuristic of common sense, which allows one to think about the dangers of surpluses that always lead to negative consequences. There are many cases for the application of this heuristic: excessive weight, excessive wealth, excessive ambition, excessive concern, etc. Almost all the evils of mankind come to the attention of this heuristic. Heuristics ‘do not provide services you are not asked for’ is related to situations: a) attitude of ingratitude to kindness; b) when the good intentions of others prevent a person from developing or violate his/her plans.  These are the situations when good turns evil for both parties.

Each of these heuristics operates independently. But if we analyze the situation using these four heuristics, then the probability of avoiding negative consequences in decisions made increases.

\paragraph {Situation 1.} 
The decision to get a dog K. has taken in order to realize the dream of her daughter, though the head of the family wasn’t happy. At the very beginning, everything was going well: the girl played with the dog, trained, fed, took the dog for a walk. Then, the girl got tired of getting up early in the morning and K. had to walk a dog herself, more than that, extra expenses appeared due to additional taxes, special food, veterinary, medicine and necessary accessories. It became clear that the dog was not a toy, but another member of the family. Since K. was the only one to take care of the puppy, she hadn’t got enough time for some things to which she was accustomed to: morning exercises, meetings with friends in the cafe and other little things. One more problem emerged: where to leave the dog when the family goes on vacation. Finally, despite the affection, the dog was given to distant relatives.

Before getting a dog, one should really assess the needs and capabilities of the family. None should succumb to positive emotions; it’s worth thinking about what and who will pay for such spontaneity (do not lie to yourself). If the dog will bring long-term joy, improve the atmosphere in the family, become a member of the family and all family members will take care of the dog, this is a good option (useful is that what leads to development). And very few people think of complex emotion, such as grief over the death of the pet, which is inevitable, because they live much shorter lives than people do (do not lie to yourself).  
\paragraph {Situation 2.} 
J. devoted too much time to work and career, pushing the family, friends and even own health to the sideline. Until recently, J. lived in the following mode: wake up, go to work, come home and go to sleep, and the next day got up to do the same things again. Looking back, J. cannot remember anything important and significant. (excess is always expensive) These years were only an endless race for the goal, which seemed to be the most important. (do not lie to yourself) He spent a lot of time to understand what was really important for him - own health, good relations with close people and personal growth. He realized that feeling tired and exhausted by life when you are 35 is not what he needs. (useful is that what leads to development)   

\subsection{Complex Heuristics}    
The study of simple heuristics irrespective of characteristics and actions of people who apply them gives many researchers wrong idea of their effectiveness. Many scientists underline different directions for application of simple heuristics with due regard to additional information. \cite{Gi11,Hu05,Ro02}.

Unlike simple heuristics, complex heuristics not only direct the actions, but also help to foresee every possible outcome. We define complex real-life heuristics as rules which disclose the cause-and-effect relationship between personality, actions of an individual and the effects of such actions. It follows that the special significance of complex heuristics is that they reveal the cause-effect relationships between the personality (morality, thinking, health, willpower) and the outcomes of their actions. For example, the lower the moral level of the person, the less gratitude he or she shows for the other people’s help. This heuristic was used by Nikos Kazantzakis in his book The Last Temptation of Christ: ‘They’ve grown rabid: they want your death.’ - ‘What have I done to them?’ - ‘You sought their good, their salvation. How can they ever pardon you for that!’

If we consider heuristics from the point of view of knowledge, it’s clear that complex heuristics are part of our social heritage. They are stable and almost time or nationality independent. That is why we understand arts of literature, movies and pictures created in different times and countries. Life heuristics can be found in various sources, including modern books on economics, philosophy, management, psychology.  The more often the heuristic is met, the more universal, reliable and efficient it is.  The use of universal heuristics contributes to understanding of many issues associated with it!  As Niccolò Machiavelli wisely indicated in The Prince: ‘…in all cities and all peoples there still exist, and have always existed, the same desires and passions. But since these matters are neglected or not understood ... the result is that the same problems always exist in every era.’ It is worth noting that apart from common sense heuristics, based on experience, there are scientifically-grounded heuristics. For example, “with the development of personality, the importance of material goods declines” (a hierarchy of needs by A. Maslow) or "the higher the income of a person is, the less motivation each subsequent unit of payment has (the marginal utility by G. Gossen).

It follows that the level of common sense depends on the quality and number of simple and complex heuristics learned by a person. The opposite is also true: the higher level of knowledge and common sense a person has, the more understandable and useful the heuristics seem to him. Minda \cite{Mi15} in his research confirms that ‘the more extensive the person’s knowledge is, the more likely the heuristic will be to provide the correct answer or optimal decision. Heuristics are faster than working through a problem, and they are usually correct.’ The scientist believes, that the heuristic works really well only if you have a rich knowledge base to draw upon when making the inferences and generating solutions.

Complex heuristics are useful for solving the majority of life problems. Their stable nature shapes people's reasoning. \cite{Va18} This is especially valuable if we lack the necessary information. For example, actions of particular people allow us to assess their personal characteristics. This attribute of complex heuristics is proved by the findings of Russell and Norvig, who declare that ‘knowledge of action outcomes enables problem-solving agents to perform well in complex environments.' \cite[p.~194]{Ru03} 

One can tell a lot about a person if one knows the outcomes of his/her actions and his/her personality traits. Using complex heuristics, reflecting the cause-effect relations between the characteristics of a person and the outcomes of his/her actions, it is possible to develop a hypothesis and then test it with due regard to the facts and weak signals.  For example, M. treats people with contempt (‘they “trust me”. Dumb fucks’ \cite{Va10}) (the heuristics as follows applies: arrogance always indicates a low level of moral development). When he was offered cooperation, he stole the idea and having finalized it, launched his own project (here we receive the signal which is proved by the heuristics).  The heuristic ‘the lower the level of morality of a person is, the less civilized methods he/she is ready to use to defend his/her interests’ gives us grounds to assume that when dealing with such a person it is vital to bear in mind that he/she will not look to the interests of others. 

The facts that we know about Mark Zuckerberg and the history of Facebook completely correspond to this heuristic. After Eduardo Saverin, with whom they started business, proved to be less useful than originally thought, Mark got rid of him without any regret. (‘We basically now need to sign over our intellectual property to a new company and just take the lawsuit…I'm just going to cut him out and then settle with him.’ \cite{Ca12}). Today, everyone knows that Facebook has breached ethical principles, having disclosed personal data of its users. But, a social network is a business that is closely connected with the trust of a large number of people in the system and its founder.  The lack of trust inevitably results in decrease of economic efficiency of the business. Maybe this is the reason why Ars Technica \cite{Ga19} ‘awarded’ Facebook the first place in the so-called Deathwatch, a list of large companies that won’t survive till 2020.

Knowing heuristics and understanding how heuristics work can help to gain life experience without negative effects. This is especially important because in real life the combination of high morale with high order common sense always focused on a win-win strategy as opposed to win-lost used by low-morale people.  This can be illustrated by parable from Foundation by Isaac Asimov: ‘A horse having a wolf as a powerful and dangerous enemy lived in constant fear of his life. Being driven to desperation, it occurred to him to seek a strong ally. Whereupon he approached a man, and offered an alliance, pointing out that the wolf was likewise an enemy of the man. The man accepted the partnership at once and offered to kill the wolf immediately, if his new partner would only co-operate by placing his greater speed at the man's disposal. The horse was willing, and allowed the man to place bridle and saddle upon him. The man mounted, hunted down the wolf, and killed him.  The horse, joyful and relieved, thanked the man, and said: “Now that our enemy is dead, remove your bridle and saddle and restore my freedom.” Whereupon the man laughed loudly and replied, “Never!” and applied the spurs with a will.’ Is the person in this parable a person with high moral values? No! Does his behavior correspond to his understanding of common sense? Yes!

\subsection{System of Heuristics}    
In the course of research it has become obvious, that a system of common sense heuristics arranged in a certain way establishes a reliable database of real-life knowledge.  Therefore we classified heuristics against the following areas: heuristics for relationship; heuristics for other life situations; heuristics for building personal models; heuristics to explain the obtained outcomes. (See Figure 1)

\begin{figure} [h]
  \centering
  \includegraphics[width=10 cm]{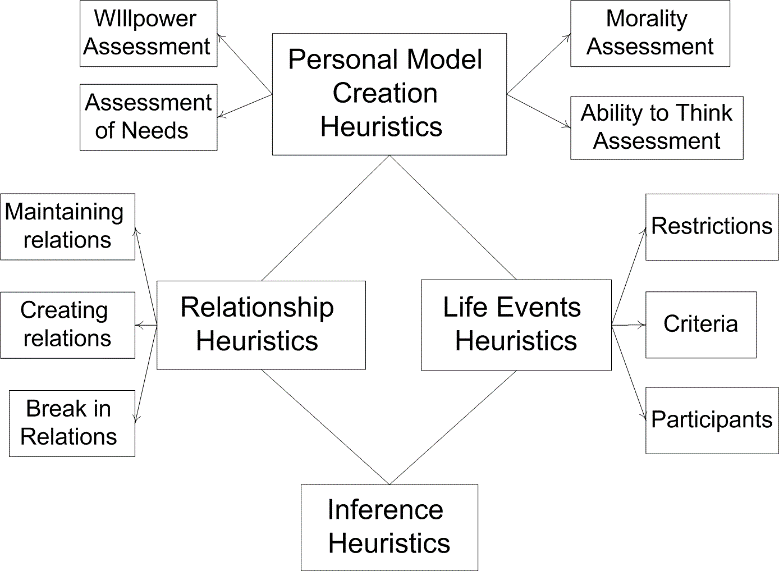}
  \caption{Heuristics Knowledge Base}
\end{figure} 

It should be underlined that the search of adequate heuristics is a hard job. Searching for universal real-life heuristics, our team has investigated 400 000 data entries. If we met the same rule in a number of sources, searched purposefully and did not find a condition under which that rule did not run, we added such a rule to the knowledge base.  As a result, more than seventy heuristics have been found that can form the basis of the life knowledge database in view of interpersonal interactions.

This can be exemplified by a heuristic as follows: ‘The less restrained one sounds, the more problems he/she faces’ heuristic connects unrestraint with its negative impacts, i.e. hostility, negative emotions, quarrels. Such common sense rules can be found in this or that form in various sources.

\begin{itemize}
\item ‘Silence is a true friend who never betrays’ (Confucius);
\item ‘Discretion of speech is more than eloquence’ (Francis Bacon);
\item ‘It is better to keep your mouth closed and let people think you are a fool than to open it and remove all doubt’ (Mark Twain);
\item ‘Long tongues sow enmity among neighbors’ (Walter Scott);
\item ‘While it is the essence of wisdom on the part of a minister of state to speak little, it is also wise to listen a great deal’ (Cardinal Richelieu);
\item ‘When there are many words, transgression is unavoidable, but he who restrains his lips is wise’ (Solomon's Proverbs);
\item ‘The word is silver, and silence is gold’ (Russian proverb);
\item ‘Listen and stay silent’ (Sophocles);
\item ‘The best habit is to keep your mouth shut’ (Arabic proverb);
\item ‘Nature has given us two ears, two eyes, and but one tongue-to the end that we should hear and see more than we speak’ (Socrates);
\item ‘He who talks a lot often fails’ (Lao Tzu);
\item ‘A friend comes from afar, the enemy comes out of your mouth’ (African proverb).
\end{itemize}

\subsection{Heuristic Algorithms} 
If we take a closer look at what people do, we’ll realize that everything people do is performed by certain algorithms. We perform our work in accordance with the instructions, play the notes, read according to the rules and cook using recipes. Many algorithms are stored in memory and applied automatically. But this does not relate to all areas of our lives, the most complex of which is decision-making in view of person-to-person interactions. Even those who are aware of the decision-making theory in practice seldom go through all stages of decision-making process, i.e. problem-formulation or target-setting; gathering the information; target clarification; selection of the parameters for alternatives evaluation; generation of alternatives; alternatives evaluation; assessment of consequences.

Unfortunately, it is impossible to effectively implement this process not knowing what actions should be performed at each stage. Some adequate solutions or, in other words, experience is needed. And what if one lacks such experience or hasn’t got enough experience? Or, being in a tense situation, a person feels heavily burdened? Or the decision shall be taken when one is pressed for time? Then it's time to apply heuristic algorithms. The knowledge base of heuristics allow for the development of common sense algorithm, backing up decision-making in view of relations and challenging situations.  Heuristic algorithms are logically grounded routes proposed to solve similar problems based on a finite set of universal rules and restrictions. They help to identify the problem and constraints affecting the solution. Many decisions taken by people can be described using such algorithms. The question is, whether in their daily interactions, people are guided by certain predictability derived from their own previous experience. According to Simon \cite{Si57} the choice of behavior patterns does not depend on a particular event, but it is predetermined by a set of rules formed under the influence of external environment. Thus, people use a limited number of heuristics and this reduces the complex tasks of assessing probabilities and predicting values to simpler judgment operations. In other words, in daily life people think using the rules learnt and act using the algorithms that they have developed independently or acquired externally. 

Presented in the form of decision trees, they ensure prompt analysis of life events and the chance to see the possible consequences of the decisions being taken. The logic component of algorithms is based on understanding of the constraints that prevent us from achieving the set goals.  The fact is that some people do not see the existing constraints, while others side-track a problem instead of eliminating them. But constraints have the most awesome feature ever – they can’t just disappear, and if one ignores them, he/she shall be sure that penalty is just a matter of time and that such penalty will inevitably result in great loss. Constraints can be external (financial, time, material) and internal (health, thinking, morality, willpower). For example, there are a number of algorithms that allow you to predict the outcomes of relationships. They are based on the basic principles of relations that are grounded on the resources that one has and the other needs. For a successful relationship, it is important that we both give and receive what we really need.

Depending on user interests and characteristics, the branch for the situation analysis and forecasting shall be determined at each stage of obtaining information. Cause-and-effect relationships are the branches for the situation analysis. Thus, there is a heuristic: if the implementation of your solution depends on others, the probability of achieving the desired result is significantly reduced.  Therefore, at this stage of the algorithm it is necessary to understand how to minimize the participation of others, to evaluate them and one’s own ability to implement the solution. For this, each basic algorithm has personal models that allow us to assess people by certain characteristics (self-evaluation and evaluation of partners). (See Figure 2)

\begin{figure} [ht]
  \centering
  \includegraphics[width=10 cm]{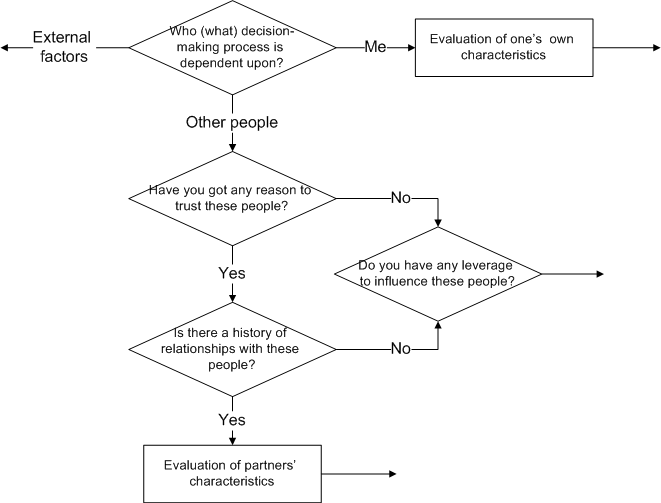}
  \caption{Heuristic Decision-Making Algorithm (Fragment)}
\end{figure} 

\subsection{Personal Models}
Building personal models is an effective method of obtaining knowledge about people and minimizing mistakes in relationships. Such models have significant value for the evaluation of individual's real interests, selection of employees, organization of a workable team, simulation of a digital copy of a person, as well as for practical psychology and pedagogy. For machines which shall possess common sense it is vital to have adequate models of people and their possible interactions. The most important personality traits, in terms of interpersonal interactions, are as follows: health, common sense (smart or silly), morality and willpower. These are the traits people pay attention to when they intuitively assess each other. Since such traits remain unchanged over a fairly long period, they are the best to build personal models on.  

Currently, there are a number of methods in the public domain, based on which it is possible to obtain evaluation data on a person. Some of them are based on methods that require personal communication with a particular specialist, most often psychologists, while others look for big data dependencies. We will not analyze these methods in detail. Each of them has its own advantages and disadvantages. We’d like to underline the following. First, many of them are misused and cannot answer some important questions. For example, one has found out that a person possesses the qualities of a leader. But we will not get the answer to the question whether this leader will get the best out of our team or will get our team away from our business. As a rule, we’ll find out the truth post factum.  Second, virtually all methods involve highly paid experts and a long evaluation process. Third, the objectivity of the information obtained in a personal survey is of questionable validity. Direct questions are almost always aimed at zero level of reflection. This can be exemplified by the questions as follows: ‘Do you agree that most people are kind? Describe a kind person. What personality traits does he/she have? To whom one must show decency and to whom not?’ Fourth, there is a difficulty in applying the obtained information, as decoding of such information often requires the subjective evaluation by highly qualified specialists. And on top of that, from the point of view of reliability, even the most widespread methods for assessing personality traits do not have sufficient scientific evidence. \cite{Si85,Mc89,Pi93,Ca77}

Personality traits are rather complex, it follows that the methods employed for their assessment shall take into account such complexity. For example, Kohlberg's method of dilemma discussion \cite{Co87} is used to assess morality. The method deals with telling stories that involve moral dilemmas and identification of moral standards of the subjects based on their attitude to these moral dilemmas and the reasons given for the decision. The disadvantages of such method are that the moral dilemmas are always accompanied with a) great uncertainty regarding the conditions under which a real decision is made and b) the researcher’s subjective view of morality. We’ve decided to employ this method for personality traits self-assessment, omitting its disadvantages to the extent reasonably practical. To achieve this, we instead of moral dilemmas used common sense observations on the actions of people to disclose their key character traits. The bipolar discrete scales were used for assessment. This approach allowed us to compare the estimates of personality traits of one and the same person against different periods of time, or of different people against the same period of time.

In supplementary material, we present the results of the experiment aimed at finding the most reliable approach to self-assessment of personality traits. We assessed fairness, which, along with responsibility and arrogance, represents morale values of a person. By fairness we understood the same judgments on one’s own action and those of other people. As practice shows, the fairer the person is, the higher moral values he/she has. But the more arrogant the person is, the lower his morale is. To evaluate fairness we’ve divided 207 people into 15 groups, composed of the individuals who knew each other (friends, colleagues, relatives). We rejected those who knew the other group members for less than a year.  The groups with less than 3 members did not participate in the study. As a result, 82 people remained; they were divided into 10 groups of 5 to 12 members. 

At the first stage, the group members assessed fairness of each other with due regard to their actions in real-life situations and mark it on a scale of 1 to 10. The integral estimate for the group was derived as an arithmetic average of individual estimates. Group assessment seems to be the most reliable, because it allows for integrated assessment of a person. Therefore, the group assessment was used as a benchmark. At the second stage, each member of the group assessed his own fairness against the same scale (direct self-evaluation). At the third stage, each indicated the degree of their readiness to be guided by certain observations on morality (indirect self- evaluation) (on a scale from 1 to 10). At this stage we used the criterion of pessimism to obtain integral indicators characterizing fairness (1).

\begin{equation}
a_i=min(x_{ij})
\end{equation}

x is the estimate of the attitude to a particular statement, j is the number of statements characterizing a person’s character traits. 

We’ve chosen the criterion of pessimism because the level of human morality and common sense is determined precisely by the level of the most non-moral actions and those that cannot be explained in terms of common sense.

For evaluation were proposed several statements: 

A) ‘To control a person one should learn about his/her weaknesses.’ It follows that a person with a high level of fairness will not focus his attention on the weaknesses of the other people to satisfy his/her own needs. 

B) ‘The end justifies the means.’ It follows that a person with a high level of fairness will not use wrong means to achieve his/her own goals.

The findings show that a direct assessment of one’s own level of fairness deviated from the group’s average score by 1.76 points. At the same time, a comparison of the estimates obtained by the indirect method and the group method showed an average deviation of 0.49. It follows that the indirect method is relatively reliable and can be applied for self-assessment of personality traits. For sure, the heuristic evaluation is probabilistic, but it allows us to quantify the assumption of the development of certain characteristics of a person.

Taking into consideration the findings of Haidt and Bjorklund \cite{Ha08} specifying that people usually cannot explain why they think something is right or wrong, or why they have done what they have done, it should be noted that an indirect assessment is extremely useful to predict the outcomes of possible interactions.  Moreover, no third party is involved in such assessment, so it is faster and cheaper than the group one.

It is worth to mention that having defined quantitative level of personality traits (morality, common sense, willpower), we may understand the behavior of people in this or that situation, and what is more important, we can compare them to predict interaction efficiency. Complex heuristics that reflect causal relationships between character traits of a person and his/her actions can contribute to the inference of the findings. For example, the lower the moral level of the person is, the less civilized methods he/she is ready to use to achieve the goal.

\section{Practical Application of Common Sense Models of Interpersonal Interactions}
\subsection{Assessment of Potential Trust }
Trust is the basis around which all human relationships revolve. In his academy publication Fukuyama \cite{Fu95} wrote  ‘The expectation that arises within a community of regular, honest, and cooperative behavior, based on commonly shared norms, on the part of other members of that community’.  Sztompka \cite{Sz99} offers another definition, which corresponds to the previous one. He defines trust as ‘a bet about the future contingent actions of others’. Trusting a doctor, for example, we believe that he will not cause harm intentionally, we expect him to follow the Hippocratic Oath and generally accepted professional medical standards. 

McKnight and Chervany \cite{Mc96} have identified the key personality traits that contribute to the emergence of trust. They highlight competence, readiness to go through the effort, honesty, and predictability of a positive result and behavior. Though scientists identify another set of personality traits necessary for the emergence of trust: predictability, conscientiousness, conformity of actions to the commitments, openness, information presentation, respect and attention. Fukuyama \cite{Fu95} also argues that trust arises when a community shares moral values and as a result can rely on regular and honest behaviour. 

According to Marcus Tullius Cicero: ‘Trust reposed in us can be established by two qualities, that is, if people come to believe that we have acquired prudence allied with justice. We put trust in those whom we regard as more perceptive than ourselves, who we believe can anticipate future events, and who at a crucial point in some action we think can cope with the situation by adopting a plan to meet the emergency – for this is the prudence which men account to be useful and genuine. As for men of justice, in other words, ‘good men’, trust in them depends on their having no suspicion of deceit and injustice in their make-up.’ \cite{Ci23}

The practice suggests that minimal indicators of trust are high moral values encompassing predictability, honesty, impartiality (also called evenhandedness or fair-mindedness), and commonsense reasoning abilities. The economic value of trust will be easier to understand if we imagine the world without trust. This is what Arrow, the Nobel Laureate in economics, says: ‘Now trust has a very important pragmatic value, if nothing else. Trust is an important lubricant of a social system. It is extremely efficient; it saves a lot of trouble to have a fair degree of reliance on other people's word. Trust is a commodity, it has real economic and practical value, it increases the efficiency of the system, enables you to produce more goods or more of whatever values you hold in high esteem. Unfortunately this is not a commodity which can be bought very easily.’ \cite{Ar74}

‘Who’s the person you trust? Jared? Who can talk you through this stuff before you decided to act on it?’ ‘Well,’ said the president, ‘you won’t like the answer, but the answer is me. Me. I talk to myself.’ \cite[p.~41]{Wo18}

Operating expenses, which are not payable in high-trust relationships, could increase because of distrust.  The role of control significantly increases as well, since in its absence people are less likely to take into account the interests of others. Control, as a rule, is carried out through confidents or authorized organizations. It follows that the category ‘trust’ is always present in this chain.  At the same time, risk of loss depends on two factors: 1) whose interests map closely to the construction of the control system and 2) technologies used to obtain and perform analysis of the data. For example, economists often refer others to some relative numbers in view of objectives set for them. But to verify the computations is almost impossible for this or that reason (different understanding of terms, peculiarities of legislation, methodology).  It follows that other objective methods of control are necessary to ensure the results close to reality. For example, according to official statistics, China GDP growth reached 6.6 pct  in 2018. \cite{NBSC} Shall we trust this data being aware about statistics double-dealing? To verify the objectivity of this figure, we have to involve control methods that are rather expensive. For example, to determine GDP growth close to the real one, experts shall from satellites track transportation and thermal radiation. 

But how can you measure the level of trust in specific people?  With due regard to the role of America in the evolving world order, we paid attention to the literature on US President Donald Trump’s personality. Wolff believes that the central issue of the Trump presidency, influencing every aspect of Trumpian policy and leadership is that ‘he didn’t process information in any conventional sense, — or rather, did not process it at all. … Trump did not like to make decisions, at least not ones that seemed to corner him into having to analyze a problem. … Almost all the professionals who were now set to join him were coming face to face with the fact that it appeared he knew nothing. … And he trusted his own expertise—no matter how paltry or irrelevant—more than anyone else’s. … On the most basic level, he simply could not link cause and effect.’ So, ignoring the warnings of almost everyone around him, Trump appointed Jared Kushner to top White House post. But six months later, the real danger of prosecution loomed over Kushner. His public role threatened not only him, but also the future of his family business.\cite{Wo18}

Regarding moral aspects, Wolff underlines that Trump evades discipline, is often cruel, violates ethics rules (he has refused to release his personal tax return, openly admitted that nothing is wrong if one has sexual relations with his friend’s wife). Peter Thiel, a co-founder of PayPal and a Facebook board member, gave a speech in support of Trump at the Republican National Convention in Cleveland, he absolutely was certain of Trump's sincerity, when Trump said they'd be friends for life—only never to basically hear from him again or have his calls returned. Trump simply could not let anyone get rich at his expense. According to the laws of his ecosystem, if one earns, the other certainly loses. James B. Comey, the former director of the F.B.I., last year during a Senate Intelligence Committee hearing in Washington, guided by his common sense, felt that the president was ‘looking to get something in exchange for granting my request to stay in the job.’ \cite[p.~336]{Wo18}  

It should be noted that in his activities Trump shows perseverance and tremendous stamina. The idea of confrontation, in which the stronger, purposeful, uncompromising, doing everything on a whim with precious little thought for consequences, wins, became the key one for Trump. He is a rebel, a destroyer, trying to live without taking into account existing rules and restrictions. Such people come to break down the existing institutional governance structures and to make the country develop in a different direction. Today, in fact, many structures - the media, the judicial system, special services, the military, - realize the interests of different groups of players. Therefore, the final result of Trump's actions will be determined by those, who have more power to apply their common sense.  For example, what benefits did Donald Trump gain from shutdown? Hundreds of thousands of federal workers at nine cabinet-level departments remained without pay for more than a month. We could assume that, having agreed to sign the government funding bill, he was defeated and did not achieve the declared goal - the construction of a wall along the border with Mexico.  But let's look at this fact from the point of view of a higher common sense. The goal of politicians who want to remain in power is often not the real fulfillment of promises that were made during the election period, but the desire to convince voters of their intentions to do so. Now, no one could say that Trump did not fulfill his campaign promise – at least he did his best.

Of course, in politics it is difficult to talk about high morals and equally developed common sense and will power, but state leaders should have objective criteria for assessing possible trust in each other, as well as assessing others' trust in them. Sociological surveys are often unstable and do not provide an objective picture. Based on the life heuristics, we have developed the BrainTrix matrix, according to which it is possible to measure in greater detail potential level of trust based on two basic indicators: the level of morality and the level of commonsense reasoning (See Figure 3).  Any person using this matrix will be able to measure the potential level of his trust in a partner and that of his partner in him.

\begin{figure} [ht]
  \centering
  \includegraphics[width=10 cm]{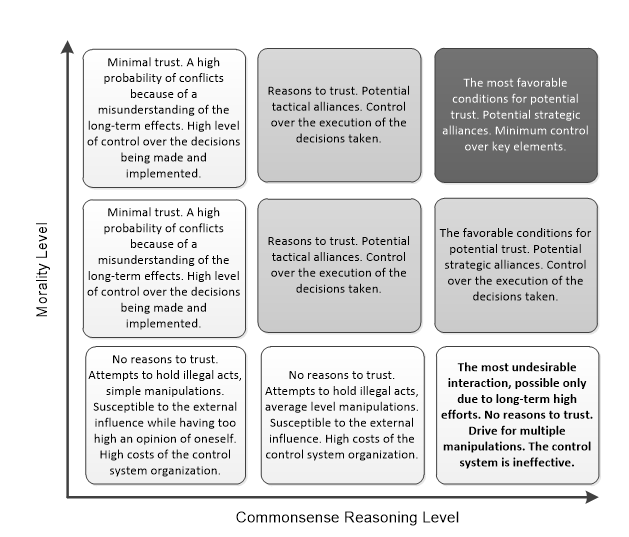}
  \caption{Potential Level of Trust in a Partner}
\end{figure} 

\subsection{Amplified Intelligence}
We have noticed that when it comes to full AI, its ability to think creatively is always implied. No one doubts that this ability is important, but for interpersonal interaction the machine shall possess reasoning skills that will ensure common sense and humanlike behaviour. The overriding concern is that machine shall help a human being to solve life problems and to take informed decisions. It follows that for successful human-machine interaction, the machine shall understand the rules and principles applied by a man. Many scientists consider this idea to be important \cite{Su17}, in particular, Bostrom and Yudkowsky, who noted that ‘an artificial intellect, by contrast, might be constituted quite differently from a human intellect yet still exhibit human‐like behavior’ \cite[p.~ 10]{Bo11}. Taking into consideration the current AI development, we believe that AI today shall be reasonably called amplified. In fact, AI is software, which possesses some commonsense knowledge, but lacks goal-setting, willpower and morality. However, moral restrictions can be imposed, if required.

The experimental prototype of BrainTrix created by us allowed for testing the concept of the machine common sense modeling with a view to interpersonal interactions. We could reasonably predict the outcomes of relationships and to assist in making tough decisions in real-life situations. Thanks to BrainTrix, we can summarize major inferences that can be drawn on the logical scheme of human-AI interaction. Such a scheme may include the aspects as follows:

a) scenario and situation-based dialogues that allow the user to interact with amplified intelligence;  

b) basic algorithms used by AI to help a user to find best solution from the point of view of common sense. Algorithms are embedded in situation-based dialogues so that the person does not notice them;

c) explanation of the essence of the actions that occur during access to the basic algorithms embedded in the situation-based dialogues;

d) personal models built by the machine based on information obtained from situation-based dialogues, and operations with this model (analysis, comparison, generation of conclusions);

e) verification of decisions against:
- correctness (compliance with moral values);
- rationality (compliance with common sense);
- effectiveness: evaluation of the decision in terms of losses/gains, health, emotions, reputation and time;

f) inferences and recommendations, illustrations and explanations.

Why such a sequence? Dialogues are necessary for the user to perceive amplified intelligence; most people engaged in concrete thinking find interaction with abstract algorithms difficult. Prototype testing confirmed our assumptions. Dialogues are written for a set of similar situations with due regard to models of individual users or created independently on the basis of artificial neural networks. Most likely, the combination of these two approaches will remain preferable for a long time. It should be noted that communication in a situation-based dialogue is personalized to increase trust of the interlocutor to the amplified intelligence. 

The problem-solving algorithm is embedded in the scenario of a dialogue. Based on incoming information, such an algorithm helps a person to identify the constraints that hinder the achievement of the goal. Depending on the ability to overcome such constraints, the user replays different scenarios of the situation. As a result, the amplified intelligence assesses the willpower, morality and reasoning capabilities of the persons involved and formulates expected outcomes with due regard to the user-defined terms and conditions.  

It is very important to verify if the decision made is effective. The end result is vital for everyone but people should give a thought to how and where this result could make itself felt. It is quite possible that solutions, inefficient in terms of money, may have other efficiency criteria. Identification of such criteria will contribute to more accurate assessment of the decisions and people’s behavior. In any case, the decision made with the help of AI is better consistent with common sense in terms of consideration of the long-term consequences.

We consider robots and voice assistants to be the most probable application for amplified intelligence. They would be in a much higher demand if they were capable of understanding people and answering like an intelligent man. Even today people apply to Siri when they need to discuss something really important. The same happens with Xiaoice developed by Microsoft. Many see Xiaoice as their partner and friend, and are willing to trust her just the way they do with their human friends. Common sense is of great need for AI. And there are enough objective reasons for this.  First, it is the systematic thinking errors caused by the peculiarities of the human brain. \cite{Li17} Tversky and Kahneman \cite{Tv74} called these errors cognitive biases.  Second, it is the inability of the average person to predict with a high degree of accuracy the outcomes of his or her actions and those of others. According to scientists, personal experience is usually not enough to respond to future situations. \cite{Ke98} As soon as the situation changes, the habitual thinking of a person can’t cope with it. Third, a person often lacks a trusted interlocutor or a wise adviser. Therefore, this is the case when external assistant, a wise amplified intelligence, shall help.  Let someone try to argue with a smart machine, the knowledge base of which is the life experience of interpersonal interactions of the whole humanity. The AI-human interaction inevitably results in training of individuals. The degree and content of such training depends on the parameters embedded in AI.

\subsection{Teaching Wisdom with the help of Heuristics }
According to scientific literature when people lack knowledge they use heuristics. But in practice people use heuristics much less frequently than researchers believe them to do. This is because the understanding of heuristics mostly depends on the real-life knowledge and the reasoning abilities of a person. It is well-known that the knowledge is divided into ‘Knowledge by Acquaintance’ and ‘Knowledge by Description'. \cite{Ru17} In this case, someone is an ‘expert’ in a narrow field and an ‘amateur’ in many others. \cite{Sc46} 

As evidenced by practice, professionals within their own spheres of competence often are not ready to solve daily life problems. This can be proved by divorce statistics. The highest divorce rate is in Portugal, where about 70 pct of marriages end in divorce. In Russia the divorce rate exceeds 60 pct. \cite{FSSS}  In Luxembourg, Spain, Denmark, Finland, almost 60 pct of marriages end in divorce, in the US the divorce rate approximates to 50 pct. \cite{Euro,NVSS}

No one will deny that commonsense reasoning is taught through life rules learnt from our mistakes made by us in different situations. The main problem is that many events or actions occur at irregular intervals, it follows that people lack skills important to decision-making when such events occur. Everyday life of many people confirm the thought of Fromm ‘the average man does not have much of an idea of what he actually is, what he wants, what he thinks about, and what he stands on’. \cite[p.~18]{Fr41}   Unfortunately, there are very few teachers of wisdom of life. This is explained by Karl Marx in his Theses on Feuerbach, where he wrote ‘the educator must himself be educated’.

Teaching knowledge for life through no negative personal experience requires serious methodological developments. The methodological basis for professional training based on heuristic algorithms was developed by Landa \cite{La76}. He introduced into psychology the ‘algorithm of mental actions’ and formed the basis for shaping thinking processes with due respect to set attributes. The fundamental difference in our case is that subject-matter for learning is real-life and human interactions heuristics. This difference is important because it significantly changes the approach to methodology. The combination of heuristics with problem solving algorithms into one system helps to cope with a lack of information and to understand the behavior of people in various situations.  In our opinion, the approach to teaching real-life thinking is as follows:

1. Identification of the most important life heuristics that a person uses daily. The common sense of these heuristics can be a starting point for evaluation of rationality of the relationship and real-life decisions against long-term consequences. There comes understanding of a) the relationship between morality, self-interest and people’s actions, decisions; b) the relationship between constraints and success of the decision taken; c) multi-criteria evaluation of the decision taken. 

2. Study of fundamental concepts of heuristics, i.e. morality, interests and needs of people, resource constraints and decision evaluation criteria. Good command of the terms and application ways ensures better understanding of a particular situation and helps to build an effective line of conduct. This is evidenced by the fact that a successful career is made faster by someone who knows the principles of the system he/she is a part of.  

3. Study of commonsense reasoning algorithms in view of interpersonal interactions. Heuristic algorithms accumulate basic mental operations (abstraction, analysis, synthesis, comparison) and life knowledge organized in the form of heuristics.  Using a system of universal algorithms one can learn to make faster and better-informed life decisions. 

4. Formation of everyday thinking skills to find the solution of a problem using universal common sense algorithms (heuristic expert system).  The student is very much in evidence of the relationship between morality, interests, constraints and action outcomes. This method allows for a unique opportunity to build "drafts of life” avoiding negative consequences and errors. Consequently, direct response to events is replaced by forecast results and proactive actions.

As a result of training, people learn to apply reasoning rules and techniques in any situation, i.e. they learn to control their own lives. With an increase in the number of studied algorithms of common sense, a person begins to create his/her own algorithms to successfully overcome life's challenges.  The symbiosis of a person and machine, possessing common sense, opens up the potential for the individualized learning. Such learning is of particular importance for educating the elite and understanding how to resist external manipulations. It’s a big deal in the era of cognitive war and attack on consciousness of people and their spiritual values.

\section{Conclusions}
We have reached the conclusion that commonsense reasoning model can be developed on several levels: first, common sense of logical physical actions and, second, common sense of interpersonal interactions. This understanding allows us to more accurately determine the modeling methods to create the most relevant models.

At the first level of modeling, due to artificial neural networks, the process develops further inward in the direction of increasing the accuracy of recognition and synthesis (images, speech, connection with bio, etc.). 

The second level, which implements the common sense of interpersonal interactions, is at the beginning of a revolutionary leap. The ability of modeling at this level will mean a substantial approximation to the creation of full artificial intelligence.

We’ve done our best to present our own vision of general process of common sense machine modeling in the field of interpersonal interactions based on a heuristic expert model. The knowledge base of such a model can be made up of simple and complex life heuristics, which are the result of collective intelligence of common sense of generations. The heuristic approach allowed us to see new possibilities for AI-human communication and interaction. We’ve tried to link knowledge representation and reasoning with the reality and outcomes of people's actions, which Russell and Norvig \cite{Ru03} believe to be central to the entire field of artificial intelligence.

Today we can model certain level of commonsense reasoning based on a set of heuristics, showing different level of rationality of a man. Amplified intelligence can become a wise interlocutor, an adviser and even a teacher of a person in difficult life situations. It can contribute to commonsense decision-making both in personal matters and in matters of professional interactions, including international relations.

\bibliographystyle {unsrt}  


\end{document}